\documentclass[]{fairmeta}

\usepackage{microtype}
\usepackage{graphicx}
\usepackage{booktabs}
\usepackage{pgfplots}
\usepackage{subcaption}
\usepackage{tikz}
\usepackage[most]{tcolorbox}
\usepackage{hyperref}
\usepackage{amsmath}
\usepackage{amssymb}
\usepackage{mathtools}
\usepackage{amsthm}
\pgfplotsset{compat=1.18} 

\title{Short Data, Long Context: Distilling Positional Knowledge in Transformers}

\author[]{Patrick Huber}
\author[]{Ernie Chang}
\author[]{Chinnadhurai Sankar}
\author[]{Rylan Conway}
\author[]{Igor Fedorov}
\author[]{Md Rifat Arefin}
\author[]{Adithya Sagar}

\affiliation[]{Meta Reality Labs}

\abstract{Extending the context window of language models typically requires expensive long-context pre-training, posing significant challenges for both training efficiency and data collection. In this paper, we present evidence that long-context retrieval capabilities can be transferred to student models through logit-based knowledge distillation, even when training exclusively on packed short-context samples within a long-context window. We provide comprehensive insights through the lens of Rotary Position Embedding (RoPE) and establish three key findings. First, consistent with prior work, we show that phase-wise RoPE scaling, which maximizes rotational spectrum utilization at each training stage, also achieves the best long-context performance in knowledge distillation setups. Second, we demonstrate that logit-based knowledge distillation can directly enable positional information transfer. Using an experimental setup with packed repeated token sequences, we trace the propagation of positional perturbations from query and key vectors through successive transformer layers to output logits, revealing that positional information systematically influences the teacher's output distribution and, in turn, the distillation signal received by the student model.
Third, our analysis uncovers structured update patterns in the query state during long-context extension, with distinct parameter spans exhibiting strong sensitivity to long-context training.}

\date{\today}
\correspondence{Patrick Huber at \email{patrickhuber@meta.com}}

\begin{document}

\maketitle

\begin{figure}
    \centering
    \includegraphics[width=0.42\linewidth]{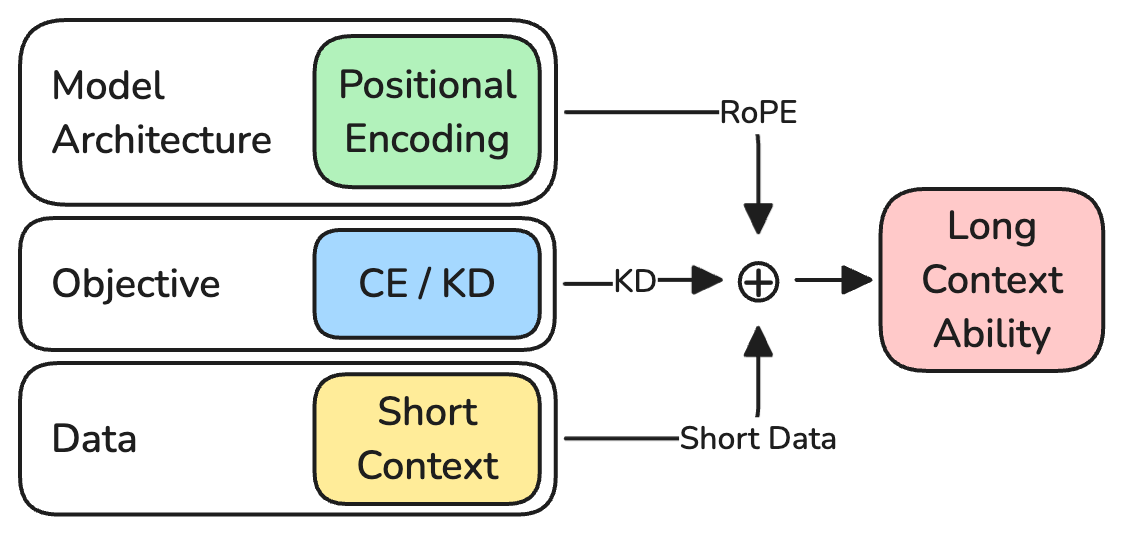}
    \caption{The three dimensions we explore in this work: RoPE embeddings, the objective function (knowledge distillation), and the underlying data. By combining these components, the student model acquires long-context abilities despite being trained only on short-context data.}
    \label{fig:overview_components}
\end{figure}

\section{Introduction}
The ability to process and reason over long sequences is increasingly critical for modern language model applications, from document summarization to multi-turn dialogues. In this work, we define \textit{long-context abilities} specifically as the capacity to accurately retrieve and attend to information across extended sequence lengths, as measured by retrieval-oriented benchmarks (Needle-in-a-Haystack, RULER)\footnote{We note that this is distinct from broader long-context reasoning competence, which may require additional capabilities beyond positional awareness.}. Training models to handle long contexts presents substantial challenges in two areas: model training, where memory requirements scale quadratically with sequence length, and data curation, where collecting large-scale, high-quality long-context training data remains difficult. While these challenges arise across all model sizes, they are particularly acute for on-device language models, where data distribution shifts between training phases can cause significant performance regressions.

In transformer architectures \citep{vaswani2017attention}, positional information is exclusively encoded in the self-attention mechanism. Among the various positional encoding schemes, Rotary Position Embedding (RoPE) \citep{su2024roformer} has emerged as the predominant approach in state-of-the-art large language models, including LLaMA \citep{touvron2023llamaopenefficientfoundation}, Gemma-3 \citep{gemmateam2025gemma3technicalreport}, Qwen-3 \citep{yang2025qwen3technicalreport}, and many others. RoPE encodes positional information by applying rotation matrices to query and key vectors, enabling models to capture both absolute and relative positions.

Building on RoPE, recent work by \citet{huber2025mobilellmprotechnicalreport} uncovered an intriguing property in knowledge distillation setups: student models distilled from long-context-capable teachers can exhibit long-context abilities, even when trained exclusively on short-context data. This finding implies that positional information encoded in the teacher's outputs can transfer to the student, enabling generalization to sequence lengths far exceeding those observed during training. This implicit transfer of positional capabilities beyond the scale of available training data raises a fundamental question about the nature of knowledge distillation and positional encodings:
\textit{How can the student model, through the distillation objective alone, acquire the ability to handle long-range dependencies?}

We hypothesize that the answer lies in the RoPE positional encoding mechanism, which perturbs query and key representations based on their sequence position. These perturbations are then propagated through the teacher's forward pass, becoming part of the knowledge distillation objective, which forces the student to match the position-dependent teacher outputs. In this way, the distillation signal carries implicit positional information, even for tokens in different packed samples that do not attend to each other.

To confirm this hypothesis, we provide a systematic investigation of implicit positional distillation, examining the interplay of RoPE embeddings, knowledge distillation objectives, and short-context training data in enabling long-context capabilities. The following research questions guide our investigation:

\textbf{RQ (1) -- RoPE Scaling for Knowledge Distillation:} What is the optimal approach for scaling the Rotary Position Embedding (RoPE) base parameter $\theta$ in knowledge distillation (KD) setups?

\textbf{RQ (2) -- Transfer of Positional Information:} Can local positional information, derived from short-context data, be leveraged for long-context learning through knowledge distillation?

\textbf{RQ (3) -- Propagation of Positional Perturbations:} While most transformer components are position-invariant, the attention query and key matrices are perturbed by RoPE positional encodings. Do these positional perturbations propagate meaningfully through the teacher model's forward pass, allowing them to serve as a training signal for the student?

\textbf{RQ (4) -- Patterns in Model Updates During Long-Context Extension:} During long-context extension via KD, can we identify specific patterns in the student model's parameter updates that correlate with positional information?

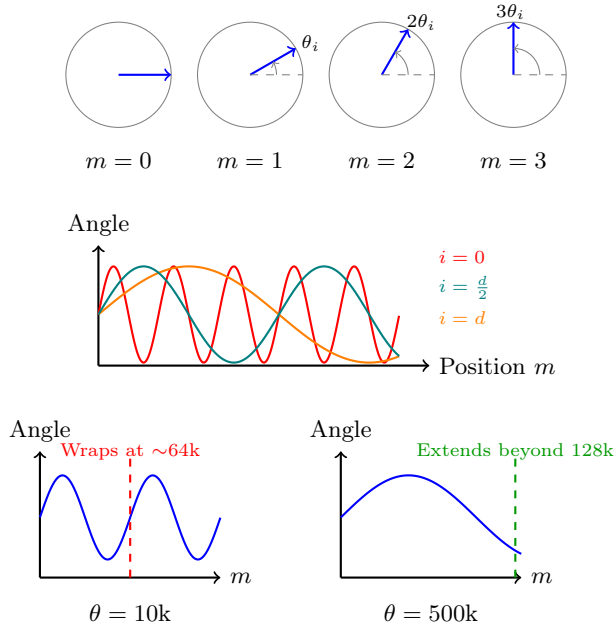
\begin{figure}[t]
\centering
    \begin{minipage}[t]{\linewidth}
    \centering
    \begin{tikzpicture}[scale=0.7]
    \draw[gray, thin] (0,0) circle (1);
    \draw[->, thick, blue] (0,0) -- ++(0:1);
    \node[below, font=\small] at (0,-1.3) {$m=0$};

    \draw[gray, thin] (2.5,0) circle (1);
    \draw[->, thick, blue] (2.5,0) -- ++(30:1);
    \draw[gray, dashed, thin] (2.5,0) -- ++(0:1);
    \draw[->, gray] (3,0.0) arc (0:30:0.5);
    \node[right, font=\scriptsize] at (3.3,0.6) {$\theta_i$};
    \node[below, font=\small] at (2.5,-1.3) {$m=1$};

    \draw[gray, thin] (5,0) circle (1);
    \draw[->, thick, blue] (5,0) -- ++(60:1);
    \draw[gray, dashed, thin] (5,0) -- ++(0:1);
    \draw[->, gray] (5.5,0) arc (0:60:0.5);
    \node[right, font=\scriptsize] at (5.3,1) {$2\theta_i$};
    \node[below, font=\small] at (5,-1.3) {$m=2$};

    \draw[gray, thin] (7.5,0) circle (1);
    \draw[->, thick, blue] (7.5,0) -- ++(90:1);
    \draw[gray, dashed, thin] (7.5,0) -- ++(0:1);
    \draw[->, gray] (8,0) arc (0:90:0.5);
    \node[left, font=\scriptsize] at (7.9,1.2) {$3\theta_i$};
    \node[below, font=\small] at (7.5,-1.3) {$m=3$};
    \end{tikzpicture}
    \end{minipage}

    \vspace{1.0em}

    \begin{minipage}[t]{\linewidth}
    \centering
    \begin{tikzpicture}[scale=0.8]
        \draw[->, thick] (0,0) -- (5.5,0) node[right, font=\small] {Position $m$};
        \draw[->, thick] (0,0) -- (0,2) node[above, font=\small] {Angle};

        \draw[red, thick, domain=0:5, samples=300] plot (\x, {0.85 + 0.8*sin(\x*360)});
        \node[right, font=\scriptsize, red] at (5.5, 1.8) {$i=0$};

        \draw[orange, thick, domain=0:5, samples=200] plot (\x, {0.85 + 0.8*sin(\x*60)});
        \node[right, font=\scriptsize, orange] at (5.5, 0.8) {$i=d$};

        \draw[teal, thick, domain=0:5, samples=200] plot (\x, {0.85 + 0.8*sin(\x*120)});
        \node[right, font=\scriptsize, teal] at (5.5, 1.3) {$i=\frac{d}{2}$};
    \end{tikzpicture}
    \label{fig:rope_intuition_b}
    \end{minipage}

    \vspace{1em}

    \begin{minipage}[t]{\linewidth}
    \centering
    \begin{tikzpicture}[scale=0.8]
        \begin{scope}[shift={(0,0)}]
            \draw[->, thick] (0,0) -- (3,0) node[right, font=\small] {$m$};
            \draw[->, thick] (0,0) -- (0,2.1) node[above, font=\small] {Angle};
            \draw[blue, thick, domain=0:3.0, samples=100] plot (\x, {1 + 0.7*sin(\x*240)});
            \node[below, font=\small] at (1.5,-0.3) {$\theta = 10\text{k}$};
            \draw[red, dashed, thick] (1.5, 0) -- (1.5, 2);
            \node[above, font=\scriptsize, red] at (1.5, 1.8) {Wraps at $\sim$64k};
        \end{scope}

        \begin{scope}[shift={(5,0)}]
            \draw[->, thick] (0,0) -- (3,0) node[right, font=\small] {$m$};
            \draw[->, thick] (0,0) -- (0,2.1) node[above, font=\small] {Angle};
            \draw[blue, thick, domain=0:3.0, samples=100] plot (\x, {1 + 0.7*sin(\x*80)});
            \node[below, font=\small] at (1.5,-0.3) {$\theta = 500\text{k}$};
            \draw[green!60!black, dashed, thick] (2.9, 0) -- (2.9, 2);
            \node[above, font=\scriptsize, green!60!black] at (2.9, 1.8) {Extends beyond 128k};
        \end{scope}
    \end{tikzpicture}
    \label{fig:rope_intuition_c}
\end{minipage}
\caption{(Top)~Each dimension pair undergoes incremental rotation as position $m$ increases, encoded by a unique rotation angle $m\theta_i$. \\
    (Center)~Different dimension pairs rotate at different frequencies: low-index pairs distinguish nearby tokens, while high-index pairs encode long-range positional relationships. \\
    (Bottom)~The base parameter $\theta$ controls how quickly angles accumulate: a smaller $\theta$ causes faster wrapping (limiting context length), while a larger $\theta$ spreads rotations across longer sequences.}
    \label{fig:rope_intuition}
\end{figure}

\section{Setup}
\label{sec:setup}
\subsection{Background: Rotary Position Embeddings}
\label{sec:rope_background}

Rotary Position Embedding (RoPE)~\citep{su2024roformer} encodes positional information by applying \textit{rotation transformations} to pairs of dimensions in the query and key vectors during self-attention. Unlike additive positional encodings that directly modify token representations, RoPE integrates position into the attention computation by rotating query and key vectors so that their dot product naturally becomes a function of their relative position, while still encoding absolute position information.

For a token at position $m$ in the sequence, RoPE groups the hidden dimension into pairs and applies a 2D rotation to each pair. The rotation matrix for the $i$-th dimension pair is defined as:

\begin{equation}
    R_m^{(i)} = \begin{pmatrix}
        \cos(m\theta_i) & -\sin(m\theta_i) \\
        \sin(m\theta_i) & \cos(m\theta_i)
    \end{pmatrix}
    \label{eq:rope_rotation}
\end{equation}

Each position in the sequence thus corresponds to a unique ``angular fingerprint'' across all dimension pairs. The base parameter $\theta$ controls the rotation frequencies across dimensions through:

\begin{equation}
    \theta_i = \theta^{-2i/d}
    \label{eq:theta_frequency}
\end{equation}

where $d$ is the hidden dimension size. This formula creates a spectrum of frequencies: low-index dimension pairs rotate quickly and distinguish nearby tokens, while high-index pairs rotate slowly and encode long-range positional relationships. A larger base $\theta$ ``slows down'' all rotations, allowing the model to distinguish positions across longer sequences before angles ``wrap around'' (i.e., the sinusoidal oscillation repeats). As a result, scaling $\theta$ is a crucial component of long-context extensions, as it expands the rotational spectrum and reduces positional ambiguity. Figure~\ref{fig:rope_intuition} provides a schematic visualization of these RoPE components.

\subsection{Models}
In our knowledge distillation training setup, we use the Llama-4 Scout frontier model \citep{meta2024llama4} as the teacher to guide the pre-training of our 1 billion parameter student model, following the approach of \citet{huber2025mobilellmprotechnicalreport} with logit-based KD. Despite the fact that Llama-4 Scout uses an iRoPE approach, interleaving RoPE and NoPE layers, our student model uses RoPE in every layer. More detailed hyperparameters for both the student and teacher are listed in Table~\ref{tab:models}.

\subsection{Training Stages}

\begin{table}[t]
\centering
\scalebox{1}{
\begin{tabular}{@{}lll@{}}
\toprule
 & Student & Teacher \\
\midrule
Model & Random Init & Llama-4 Scout \\
Layers & 30 & 48 \\
Attn Heads & 20 & 40 \\
KV Heads & 4 & 8 \\
Dimension & 1,280 & 5,120 \\
Hidden Dim & 6,144 & 16,384 \\
Vocab Size & 202,048 & 202,048 \\
Feed Forward & Dense & 16 Experts \\
Context Length & 128k tokens & 10M tokens \\
Total Params & 1.1B & 17B / 109B \\
RoPE Flavor & RoPE & iRoPE \\
\bottomrule
\end{tabular}}
\caption{Student and Teacher Model Specifications}
\label{tab:models}
\end{table}

\begin{table}[ht]
    \centering
    \begin{tabular}{|c|c|c|c|c|c|c|c|c|c|c|}
        \hline
        \textbf{Percentile} & 10 & 20 & 30 & 40 & 50 & 60 & 70 & 80 & 90 & 99 \\
        \hline
        \textbf{Word Count per Sample} & 154 & 220 & 293 & 375 & 466 & 571 & 710 & 935 & 1,441 & 4,955 \\
        \hline
    \end{tabular}
    \caption{Word count percentiles of our pre-training dataset.}
    \label{tab:wordcount_percentiles}
\end{table}

Our training procedure consists of two phases, following the standard practice of progressive context length extension:

\textbf{Stage 1 (Short Context):} In the initial pre-training phase, we train the model for 128,000 steps over 100 billion tokens. Following common practice, we use a short context window of 2,048 tokens per sequence in this stage, focusing on foundational language understanding.

\textbf{Stage 2 (Long Context):} In the second phase, we continue pre-training the model for an additional 20,000 steps over 10 billion tokens. In this stage, we expand the context window to 128,000 tokens, training the model to handle longer-range dependencies. Critically, even in this long-context phase, the underlying training data remains short-context (see Table~\ref{tab:wordcount_percentiles} for the dataset's word count distribution), with multiple short documents packed into each 128k sequence. Each packed sequence consists of concatenated, independently sampled short documents separated by EOS tokens, with a self-attention mask that prevents cross-document attention. This ensures that individual samples remain contextually isolated; the only signal that spans beyond a single document boundary is the positional encoding applied at the full sequence level. We emphasize that no long-form documents (e.g., books, lengthy articles) are included in the training data at any stage.

\section{RoPE Scaling for Knowledge Distillation}
\label{sec:ablations}
The topic of RoPE scaling factors, specifically the base parameter $\theta$, has been extensively explored in the literature \citep{liu2024scalinglawsropebased}. However, two factors distinguish our setting from prior work: (1) We consider on-device sized models, which are one to two orders of magnitude smaller than the frontier-sized models typically studied. (2) In the KD setup, we must additionally consider the teacher model's $\theta$ parameter, which may play a crucial role in enabling the transfer of positional information.

\subsection{Method: Phase-wise RoPE Settings}

Given these differences, we run a set of experiments to establish the optimal RoPE configuration under our constraints. We explore three distinct setups:

\textbf{1. Consistent, Teacher-Aligned ($\theta = 500\text{k}$):} In this configuration, we set the RoPE $\theta$ parameter to $500{,}000$ across both pre-training stages, matching the teacher model's parameterization. The intuition is to keep rotations aligned between the student and teacher models throughout training.

\textbf{2. Consistent, Literature-Aligned ($\theta = 10\text{k}$):} Here, we also maintain a consistent $\theta$ across both stages, but use the standard value of $\theta = 10{,}000$ that is common in the literature.

\textbf{3. Phase-wise Scaled ($\theta: 10\text{k} \rightarrow 500\text{k}$):} In this configuration, we use the standard $\theta = 10{,}000$ during the initial short-context pre-training stage, exposing the model to larger RoPE angles during this phase. We then increase $\theta$ to $500{,}000$ during the long-context extension phase to ensure that angles do not wrap around within the maximum sequence length of 128,000 tokens. Compared to setups (1) and (2), the intuition behind this approach is to maximize the positional information available to the model at each stage, adapting the rotational spectrum to the sequence lengths encountered during training.

\begin{figure}[h]
    \centering
    \includegraphics[width=0.5\linewidth]{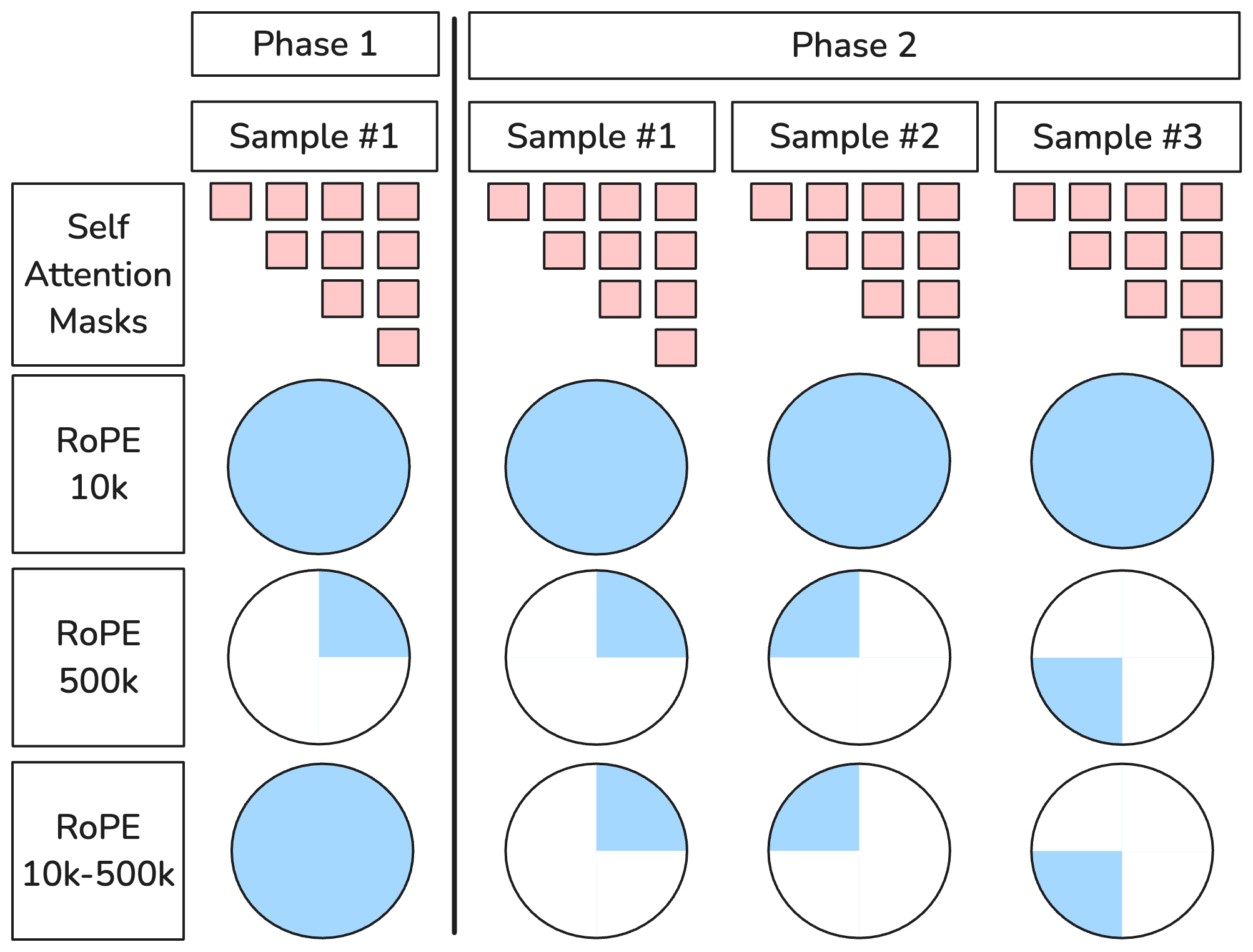}
    \caption{Schematic visualization of scaled and unscaled RoPE $\theta$ in the short-context (Phase 1) and long-context (Phase 2) pre-training stages.}
    \label{fig:overview}
\end{figure}

All three setups are visually represented in Figure \ref{fig:overview}.

\subsection{Findings}
\begin{figure*}
    \centering
    \begin{minipage}{0.48\linewidth}
        \centering
        \begin{tikzpicture}
            \begin{axis}[
                ybar,
                bar width=15pt,
                width=\linewidth,
                height=4cm,
                ymin=0, ymax=50,
                ylabel={Average Score (\%)},
                xticklabel style={font=\small},
                xtick=data,
                xticklabel style={font=\small},
                xticklabels={10k KD, 10k-500k KD, 500k KD, 10k CE, 10k-500k CE, 500k CE},
                x tick label style={rotate=30, anchor=east},
                grid=major,
            ]
            \addplot+[fill=blue!60] coordinates {(1,25.05) (2,42.99) (3,41.29) (4,14.63) (5,24.78) (6,23.36)};
            \end{axis}
        \end{tikzpicture}
    \end{minipage}
    \begin{minipage}{0.23\linewidth}
        \centering
        \begin{tikzpicture}
            \begin{axis}[
                ybar,
                bar width=9pt,
                width=.9\linewidth,
                height=3.5cm,
                ylabel={Loss},
                xtick=data,
                xticklabel style={font=\small},
                xticklabels={10k, 10k-500k, 500k},
                title={Knowledge Distilled },
                grid=major,
                x tick label style={rotate=30, anchor=east},
            ]
            \addplot+[fill=red!60] coordinates {(1,1.55677187) (2,1.5532577) (3,1.54700041)};
            \end{axis}
        \end{tikzpicture}
    \end{minipage}
    \begin{minipage}{0.23\linewidth}
        \begin{tikzpicture}
            \begin{axis}[
                ybar,
                bar width=9pt,
                width=.9\linewidth,
                height=3.5cm,
                xtick=data,
                xticklabels={10k, 10k-500k, 500k},
                grid=major,
                xticklabel style={font=\small},
                title={Cross Entropy},
                x tick label style={rotate=30, anchor=east},
            ]
            \addplot+[fill=red!60] coordinates {((1,2.49864435) (2,2.49486113) (3,2.49036956)};
            \end{axis}
        \end{tikzpicture}
    \end{minipage}
    \caption{Comparison of the average pre-training performance (left, in \%) and training loss across different $\theta$ settings (right) for KD and CE objectives.}
    \label{fig:pt_eval_and_loss}
\end{figure*}
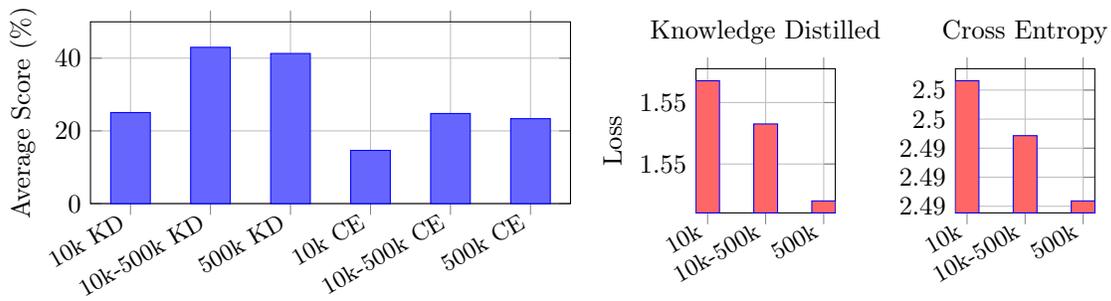

We train the student model under each of the three RoPE $\theta$ configurations using both logit-based KD and standard cross-entropy (CE) loss, and compare their performance along three dimensions: general pre-training language modeling ability, long-context ability, and training loss. The results are shown in Figures~\ref{fig:nih_ruler_eval} and~\ref{fig:pt_eval_and_loss}. We omit Phase~1 evaluation results, as long-context performance is consistently 0\% at that stage. The Phase~2 evaluations clearly address RQs~(1) and~(2):

\textbf{Finding 1: Larger $\theta$ values support long-context abilities.} Using a RoPE $\theta$ of 10,000 consistently underperforms the 500,000 and phase-wise scaled approaches, in line with the intuition illustrated in Figure~\ref{fig:rope_intuition}. The phase-wise scaled method achieves the best average performance. This answers RQ~(1): the optimal strategy is to maximize rotational spectrum utilization at each training stage by scaling $\theta$ between phases.

\textbf{Finding 2: KD improves over cross-entropy models.} KD consistently outperforms CE-trained models across all $\theta$ configurations, yielding better long-context understanding and lower training loss. This answers RQ~(2): by using a long-context-capable teacher model with logit-based KD, we can instill a meaningful degree of long-context understanding in the student, an ability that cannot emerge from the data alone in the CE setup.

Based on these two findings, we proceed with the phase-wise scaled variant ($\theta = 10{,}000$ in Phase~1, $\theta = 500{,}000$ in Phase~2) for the analyses in the following sections.\footnote{While we train our models for a significant number of tokens, they remain under-trained compared to state-of-the-art models at the 1B parameter scale.}

\begin{figure*}[t]
    \centering
    \begin{minipage}{0.59\linewidth}
        \centering
        \begin{tikzpicture}
            \begin{axis}[
                width=\linewidth,
                height=5cm,
                xlabel={NIH Needle Distance \citeyearpar{needle-in-haystack}},
                ylabel={Percentage},
                xtick=data,
                xticklabels={10k,19k,28k,37k,46k,55k,64k,74k,83k,92k,101k,110k,119k,128k},
                legend to name=sharedlegend,
                legend columns=3,
                ymin=0, ymax=110,
                x tick label style={rotate=45, anchor=east},
            ]
            \addplot+[mark=*, color=blue] coordinates {
                (1,98.33) (2,91.67) (3,95.00) (4,58.33) (5,43.33) (6,33.33) (7,28.33) (8,11.67) (9,16.67) (10,5.00) (11,5.00) (12,10.00) (13,10.00) (14,6.67)
            };
            \addlegendentry{10k $\theta$ (KD)}
            \addplot+[mark=square*, color=red] coordinates {
                (1,98.33) (2,88.33) (3,100.00) (4,100.00) (5,81.67) (6,86.67) (7,78.33) (8,71.67) (9,76.67) (10,45.00) (11,58.33) (12,48.33) (13,50.00) (14,33.33)
            };
            \addlegendentry{10k-500k $\theta$ (KD)}
            \addplot+[mark=triangle*, color=green!70!black] coordinates {
                (1,100.00) (2,100.00) (3,100.00) (4,100.00) (5,100.00) (6,100.00) (7,100.00) (8,90.00) (9,78.33) (10,55.00) (11,53.33) (12,38.33) (13,36.67) (14,38.33)
            };
            \addlegendentry{500k $\theta$ (KD)}
            \addplot+[mark=diamond*, color=purple] coordinates {
                (1,65.00) (2,30.00) (3,15.00) (4,13.33) (5,18.33) (6,21.67) (7,13.33) (8,15.00) (9,21.67) (10,20.00) (11,8.33) (12,6.67) (13,6.67) (14,5.00)
            };
            \addlegendentry{10k $\theta$ (CE)}
            \addplot+[mark=star, color=orange] coordinates {
                (1,83.33) (2,78.33) (3,78.33) (4,48.33) (5,45.00) (6,48.33) (7,25.00) (8,11.67) (9,13.33) (10,15.00) (11,6.67) (12,6.67) (13,13.33) (14,8.33)
            };
            \addlegendentry{10k-500k $\theta$ (CE)}
            \addplot+[mark=otimes*, color=brown] coordinates {
                (1,100.00) (2,98.33) (3,100.00) (4,98.33) (5,85.00) (6,76.67) (7,60.00) (8,50.00) (9,35.00) (10,20.00) (11,25.00) (12,18.33) (13,6.67) (14,8.33)
            };
            \addlegendentry{500k $\theta$ (CE)}
            \end{axis}
        \end{tikzpicture}
    \end{minipage}
    \hfill
    \begin{minipage}{0.4\linewidth}
        \centering
        \begin{tikzpicture}
            \begin{axis}[
                width=\linewidth,
                height=5cm,
                xlabel={RULER Needle Distance \citeyearpar{hsieh2024rulerwhatsrealcontext}},
                xtick=data,
                xticklabels={4k,8k,16k,32k,64k,128k},
                legend to name=sharedlegend,
                legend columns=3,
                ymin=0, ymax=65,
                x tick label style={rotate=45, anchor=east},
            ]
            \addplot+[mark=*, color=blue] coordinates {
                (1,49.22) (2,43.60) (3,36.15) (4,26.08) (5,8.55) (6,5.39)
            };
            \addlegendentry{10k $\theta$ (KD)}
            \addplot+[mark=square*, color=red] coordinates {
                (1,52.38) (2,46.89) (3,41.60) (4,37.08) (5,31.24) (6,25.60)
            };
            \addlegendentry{10k-500k $\theta$ (KD)}
            \addplot+[mark=triangle*, color=green!70!black] coordinates {
                (1,51.84) (2,45.87) (3,41.36) (4,39.13) (5,30.51) (6,14.16)
            };
            \addlegendentry{500k $\theta$ (KD)}
            \addplot+[mark=diamond*, color=purple] coordinates {
                (1,33.18) (2,25.29) (3,19.96) (4,11.87) (5,7.14) (6,2.02)
            };
            \addlegendentry{10k $\theta$ (CE)}
            \addplot+[mark=star, color=orange] coordinates {
                (1,39.94) (2,34.56) (3,33.89) (4,25.96) (5,19.27) (6,10.74)
            };
            \addlegendentry{10k-500k $\theta$ (CE)}
            \addplot+[mark=otimes*, color=brown] coordinates {
                (1,41.45) (2,26.69) (3,22.90) (4,20.25) (5,16.20) (6,12.03)
            };
            \addlegendentry{500k $\theta$ (CE)}
            \end{axis}
        \end{tikzpicture}
    \end{minipage}

    \vspace{0.5em}
    \centering
    \pgfplotslegendfromname{sharedlegend}
    \caption{Performance for different RoPE $\theta$ values and training methods (CE vs.\ KD) across Needle in a Haystack (left) and RULER (right).}
    \label{fig:nih_ruler_eval}
\end{figure*}
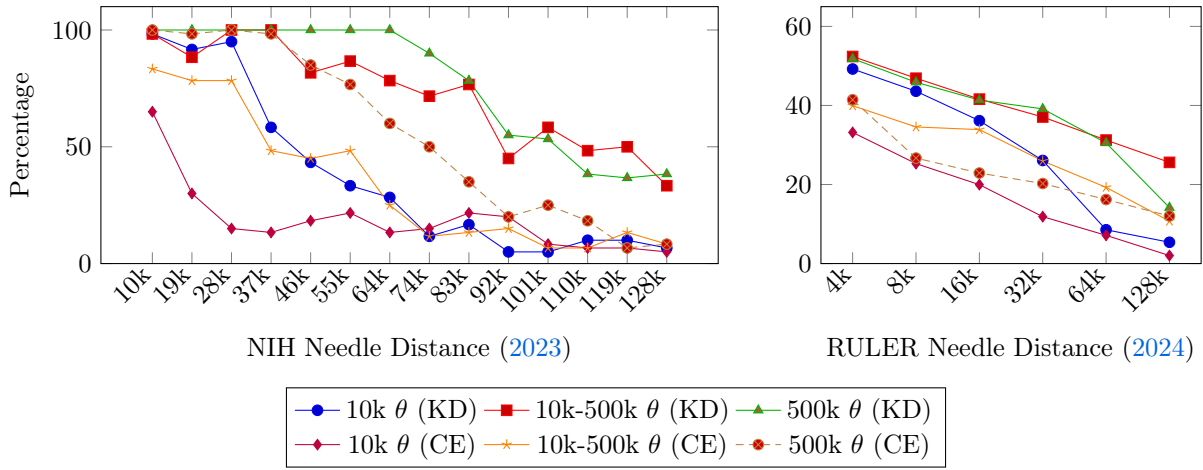

\section{Positional Perturbations in Transformer Architectures}
\label{sec:analysis}

Having established that knowledge distillation with appropriate RoPE scaling enables long-context capabilities, we now turn to investigating the mechanism underlying this transfer. Specifically, we ask how positional perturbations, introduced by RoPE at the query and key level, propagate through the teacher model's forward pass, and do they ultimately influence the output logits that serve as the distillation signal?

\subsection{Method: Isolating Positional Information}
To analyze the propagation of positional information, we design an experiment that isolates positional effects from semantic and syntactic content. We construct an input sequence by repeating a fixed block of 2,048 tokens, a BOS (beginning-of-sequence) token, 2,046 content tokens, and an EOS (end-of-sequence) token, 64 times, filling the model's full context window of 128,000 tokens. Because every ``packed'' segment is semantically and syntactically identical, any observed differences in hidden representations between repetitions can be attributed solely to positional perturbations introduced by RoPE.

During the forward pass, we systematically log hidden states at key stages: the raw input text, the tokenized sequence, the embedding layer, the query hidden state both before and after RoPE application, the final layer hidden state (post-feedforward), and the output logits from the language modeling head. For analysis, we extract the representation of the final token before each EOS, as it possesses the maximal local context within its segment.

\begin{figure*}[t]
    \centering
    \includegraphics[width=0.85\linewidth]{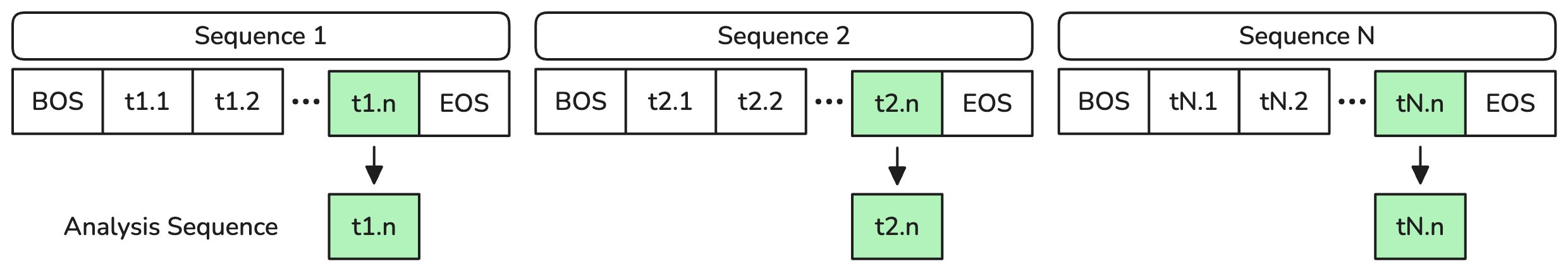}
    \caption{Schematic visualization of the experimental approach. We repeat a fixed sequence of $n=2{,}048$ tokens $N=64$ times to generate an analysis sequence of shape $(bsz, 64, dim)$, filling the complete context of $131{,}072$ tokens. By using identical content at every position, we isolate positional perturbations from semantic and syntactic differences.}
    \label{fig:experiment_schematic}
\end{figure*}

This yields an analysis tensor of shape $(bsz, 64, dim)$, where $bsz = 1$ (a single input sample), the sequence dimension contains 64 tokens each separated by 2,048 positions (representing snapshots across the full token sequence), and $dim$ is the hidden-state dimensionality. Figure~\ref{fig:experiment_schematic} illustrates this process, and Appendix~\ref{app:inputs_and_tokens} provides the input text, token sequence, and resulting analysis sequence used in our experiments.

\subsection{Findings}
\paragraph{Inputs and Embeddings}
As expected, the input tokens, vocabulary embeddings, and pre-RoPE query hidden states are position-invariant across all 64 repetitions. This validates our experimental design and confirms that any subsequent divergence in representations is attributable to positional encoding rather than input content.

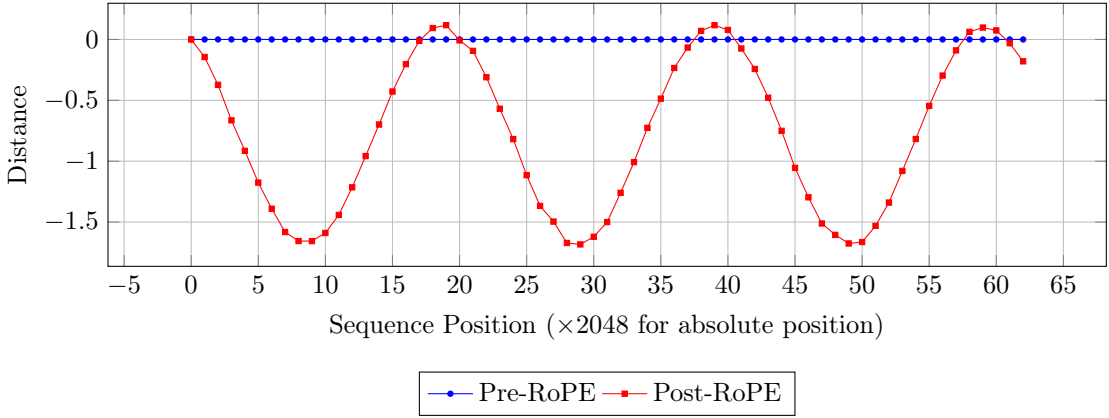
\begin{figure}[t]
    \centering
    \begin{tikzpicture}
        \begin{axis}[
            scale only axis,
            width=0.8\linewidth,
            height=3.5cm,
            xlabel={Sequence Position ($\times$2048 for absolute position)},
            ylabel={Distance},
            legend style={at={(0.5,-0.4)},anchor=north,legend columns=-1},
            grid=major,
        ]
        \addplot[
            color=blue,
            mark=*,
            mark size=1pt,
        ] table[row sep=\\]{
        0 0.0 \\
        1 -0.0 \\
        2 -0.0 \\
        3 -0.0 \\
        4 0.0 \\
        5 0.0 \\
        6 0.0 \\
        7 0.0 \\
        8 0.0 \\
        9 0.0 \\
        10 0.0 \\
        11 -0.0 \\
        12 0.0 \\
        13 -0.0 \\
        14 -0.0 \\
        15 -0.0 \\
        16 0.0 \\
        17 0.0 \\
        18 0.0 \\
        19 0.0 \\
        20 -0.0\\
        21 0.0 \\
        22 0.0 \\
        23 -0.0 \\
        24 0.0 \\
        25 0.0 \\
        26 0.0 \\
        27 -0.0 \\
        28 0.0 \\
        29 0.0 \\
        30 0.0 \\
        31 0.0 \\
        32 0.0 \\
        33 0.0 \\
        34 0.0 \\
        35 0.0 \\
        36 0.0 \\
        37 -0.0 \\
        38 0.0 \\
        39 0.0 \\
        40 0.0 \\
        41 0.0 \\
        42 0.0 \\
        43 0.0\\
        44 0.0 \\
        45 0.0 \\
        46 -0.0 \\
        47 0.0 \\
        48 -0.0 \\
        49 0.0 \\
        50 0.0 \\
        51 0.0 \\
        52 0.0 \\
        53 -0.0 \\
        54 -0.0 \\
        55 -0.0 \\
        56 0.0 \\
        57 0.0 \\
        58 0.0 \\
        59 -0.0 \\
        60 -0.0 \\
        61 -0.0 \\
        62 0.0 \\
        };
        \addlegendentry{Pre-RoPE}

        \addplot[
            color=red,
            mark=square*,
            mark size=1pt,
        ] table[row sep=\\]{
        0 0.0 \\
        1 -0.14453125 \\
        2 -0.373046875 \\
        3 -0.66455078125 \\
        4 -0.9150390625 \\
        5 -1.17578125 \\
        6 -1.390625 \\
        7 -1.58203125 \\
        8 -1.65625 \\
        9 -1.65625 \\
        10 -1.58984375 \\
        11 -1.44140625 \\
        12 -1.21484375 \\
        13 -0.9580078125 \\
        14 -0.69873046875 \\
        15 -0.427734375 \\
        16 -0.203125 \\
        17 -0.01171875 \\
        18 0.09375 \\
        19 0.1171875 \\
        20 -0.0078125 \\
        21 -0.09375 \\
        22 -0.310546875 \\
        23 -0.5693359375 \\
        24 -0.818359375 \\
        25 -1.115234375 \\
        26 -1.3671875 \\
        27 -1.49609375 \\
        28 -1.671875 \\
        29 -1.68359375 \\
        30 -1.62109375 \\
        31 -1.5 \\
        32 -1.259765625 \\
        33 -1.0078125 \\
        34 -0.72607421875 \\
        35 -0.486328125 \\
        36 -0.234375 \\
        37 -0.06640625 \\
        38 0.0703125 \\
        39 0.1171875 \\
        40 0.078125 \\
        41 -0.07421875 \\
        42 -0.2421875 \\
        43 -0.478515625 \\
        44 -0.750732421875 \\
        45 -1.0546875 \\
        46 -1.296875 \\
        47 -1.51171875 \\
        48 -1.60546875 \\
        49 -1.67578125 \\
        50 -1.6640625 \\
        51 -1.53125 \\
        52 -1.33984375 \\
        53 -1.080078125 \\
        54 -0.81787109375 \\
        55 -0.5458984375 \\
        56 -0.296875 \\
        57 -0.08984375 \\
        58 0.0625 \\
        59 0.09765625 \\
        60 0.07421875 \\
        61 -0.03125 \\
        62 -0.1796875 \\
        };
        \addlegendentry{Post-RoPE}
        \end{axis}
    \end{tikzpicture}
    \caption{Distance between position index 0 and subsequent positions for a single hidden-dimension element in the query hidden state, measured before (Pre-RoPE) and after (Post-RoPE) applying rotary positional embeddings.}
    \label{fig:abs_delta_single_dim}
\end{figure}

\paragraph{Query Vectors and RoPE Perturbations}
Figure~\ref{fig:abs_delta_single_dim} shows the distance 
\begin{equation}
    dist(A, B) = \sum_{i_1, \dots, i_{|d|}} |A_{i_1, \dots, i_{|d|}} - B_{i_1, \dots, i_{|d|}}|
\end{equation}
between the first sequence element $A$(at index~0) and each subsequent element $B$ at absolute position indices $2{,}048 \times N$ (with $N = 1, \ldots, 63$) for a single hidden-state element $d \in \text{dim}$, consistent with the schematic in Figure~\ref{fig:rope_intuition}. We compute this single-element distance both before and after applying RoPE perturbations. As expected, the pre-RoPE values are position-invariant, while the post-RoPE values follow the characteristic oscillation pattern described in Figure~\ref{fig:rope_intuition} (top).

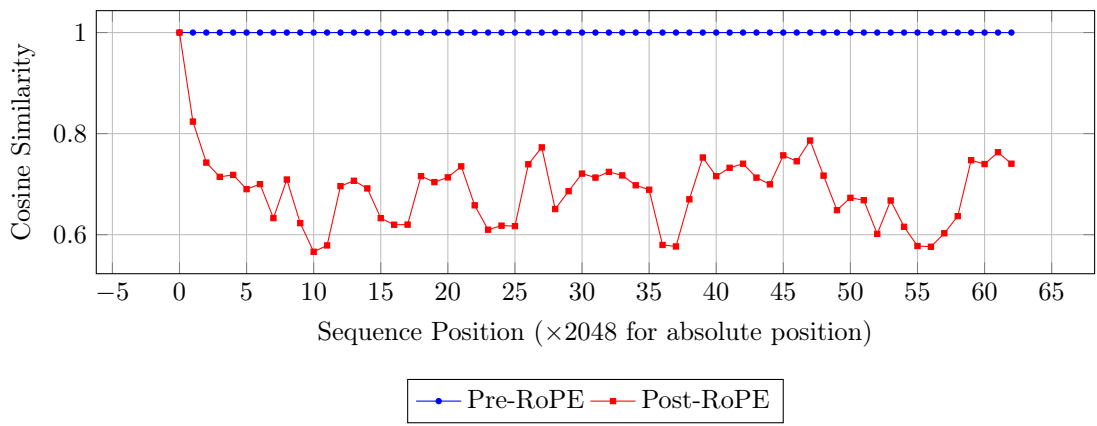
\begin{figure}[t]
    \centering
    \begin{tikzpicture}
        \begin{axis}[
            width=0.8\linewidth,
            scale only axis,
            height=3.5cm,
            xlabel={Sequence Position ($\times$2048 for absolute position)},
            ylabel={Cosine Similarity},
            legend style={at={(0.5,-0.4)},anchor=north,legend columns=-1},
            grid=major,
        ]
        \addplot[
            color=blue,
            mark=*,
            mark size=1pt,
        ] table[row sep=\\]{
        0 1.0 \\
        1 1.0 \\
        2 1.0 \\
        3 1.0 \\
        4 1.0 \\
        5 1.0 \\
        6 1.0 \\
        7 1.0 \\
        8 1.0 \\
        9 1.0 \\
        10 1.0 \\
        11 1.0 \\
        12 1.0 \\
        13 1.0 \\
        14 1.0 \\
        15 1.0 \\
        16 1.0 \\
        17 1.0 \\
        18 1.0 \\
        19 1.0 \\
        20 1.0 \\
        21 1.0 \\
        22 1.0 \\
        23 1.0 \\
        24 1.0 \\
        25 1.0 \\
        26 1.0 \\
        27 1.0 \\
        28 1.0 \\
        29 1.0 \\
        30 1.0 \\
        31 1.0 \\
        32 1.0 \\
        33 1.0 \\
        34 1.0 \\
        35 1.0 \\
        36 1.0 \\
        37 1.0 \\
        38 1.0 \\
        39 1.0 \\
        40 1.0 \\
        41 1.0 \\
        42 1.0 \\
        43 1.0 \\
        44 1.0 \\
        45 1.0 \\
        46 1.0 \\
        47 1.0 \\
        48 1.0 \\
        49 1.0 \\
        50 1.0 \\
        51 1.0 \\
        52 1.0 \\
        53 1.0 \\
        54 1.0 \\
        55 1.0 \\
        56 1.0 \\
        57 1.0 \\
        58 1.0 \\
        59 1.0 \\
        60 1.0 \\
        61 1.0 \\
        62 1.0 \\
        };
        \addlegendentry{Pre-RoPE}

        \addplot[
            color=red,
            mark=square*,
            mark size=1pt,
        ] table[row sep=\\]{
        0 1.0 \\
        1 0.8238297700881958 \\
        2 0.742685854434967 \\
        3 0.7144474983215332 \\
        4 0.718322217464447 \\
        5 0.690118670463562 \\
        6 0.7000986337661743 \\
        7 0.6328237056732178 \\
        8 0.7091251611709595 \\
        9 0.6228535771369934 \\
        10 0.566150963306427 \\
        11 0.5787219405174255 \\
        12 0.6961015462875366 \\
        13 0.7066804766654968 \\
        14 0.6915299892425537 \\
        15 0.632696270942688 \\
        16 0.6197426915168762 \\
        17 0.6199264526367188 \\
        18 0.7158781290054321 \\
        19 0.7040920853614807 \\
        20 0.7134259939193726 \\
        21 0.7351970076560974 \\
        22 0.6580577492713928 \\
        23 0.6098971366882324 \\
        24 0.6178863048553467 \\
        25 0.6169783473014832 \\
        26 0.7393547892570496 \\
        27 0.7727404236793518 \\
        28 0.6507096886634827 \\
        29 0.6860789656639099 \\
        30 0.7208945155143738 \\
        31 0.7129859924316406 \\
        32 0.7243263125419617 \\
        33 0.7172573208808899 \\
        34 0.6976965665817261 \\
        35 0.6890292167663574 \\
        36 0.5796555876731873 \\
        37 0.5767112374305725 \\
        38 0.6699525713920593 \\
        39 0.7526858448982239 \\
        40 0.7159018516540527 \\
        41 0.7324123978614807 \\
        42 0.7404695153236389 \\
        43 0.7129471898078918 \\
        44 0.6996617913246155 \\
        45 0.7569310665130615 \\
        46 0.7452630996704102 \\
        47 0.7863373756408691 \\
        48 0.7168940305709839 \\
        49 0.6485151648521423 \\
        50 0.6730320453643799 \\
        51 0.6682243347167969 \\
        52 0.6015316843986511 \\
        53 0.6676405668258667 \\
        54 0.6156322360038757 \\
        55 0.5775856971740723 \\
        56 0.5762068629264832 \\
        57 0.6027426719665527 \\
        58 0.6367232799530029 \\
        59 0.7473131418228149 \\
        60 0.7397805452346802 \\
        61 0.7631404399871826 \\
        62 0.7405275106430054 \\
        };
        \addlegendentry{Post-RoPE}
        \end{axis}
    \end{tikzpicture}
    \caption{Cosine similarity between the query hidden state at position index 0 and subsequent positions, measured before (Pre-RoPE) and after (Post-RoPE) applying rotary positional embeddings.}
    \label{fig:cosine_query}
\end{figure}

Moving beyond a single element to the full hidden state $H$, Figure~\ref{fig:cosine_query} shows the cosine similarity between complete hidden states at different positions, both before and after RoPE application. The pre-RoPE cosine similarity remains at 1.0 across all positions, confirming position invariance. In contrast, the post-RoPE cosine similarity decreases substantially, indicating that positional perturbations differentiate the representations of otherwise identical tokens. Notably, the similarity does not decrease monotonically with distance; instead, it exhibits a complex, non-monotonic pattern resulting from the superposition of multiple oscillatory components across dimension pairs, as illustrated in Figure~\ref{fig:rope_intuition} (center).

\begin{figure}[t]
    \centering
    \includegraphics[width=0.5\linewidth]{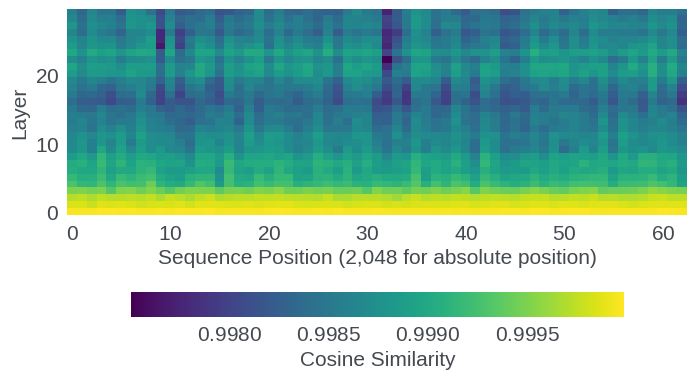}
    \caption{Per-layer cosine similarity between position index 0 and subsequent position indices (offset by 2,048 tokens each). Each curve represents a different layer, showing progressive amplification of positional divergence through the model.}
    \label{fig:per_layer}
\end{figure}

\paragraph{Hidden States Across Model Layers}
Tracking these effects across layers (Figure~\ref{fig:per_layer}), we find that the divergence induced by RoPE is amplified as representations propagate through the model. The representations of the same token at different positions become progressively more distinct with each layer, underscoring the compounding effect of positional encoding in deep transformer stacks. Interestingly, the difference in cosine similarity between positional indices (along the x-axis in Figure~\ref{fig:per_layer}) is again non-monotonic, consistent with the pattern observed in Figure~\ref{fig:cosine_query}. We believe that this behavior arises because the superposition of multiple oscillatory components at different frequencies creates a complex interference pattern that does not exhibit a simple distance-dependent decay.

\begin{figure}[t]
    \centering
    \includegraphics[width=\linewidth]{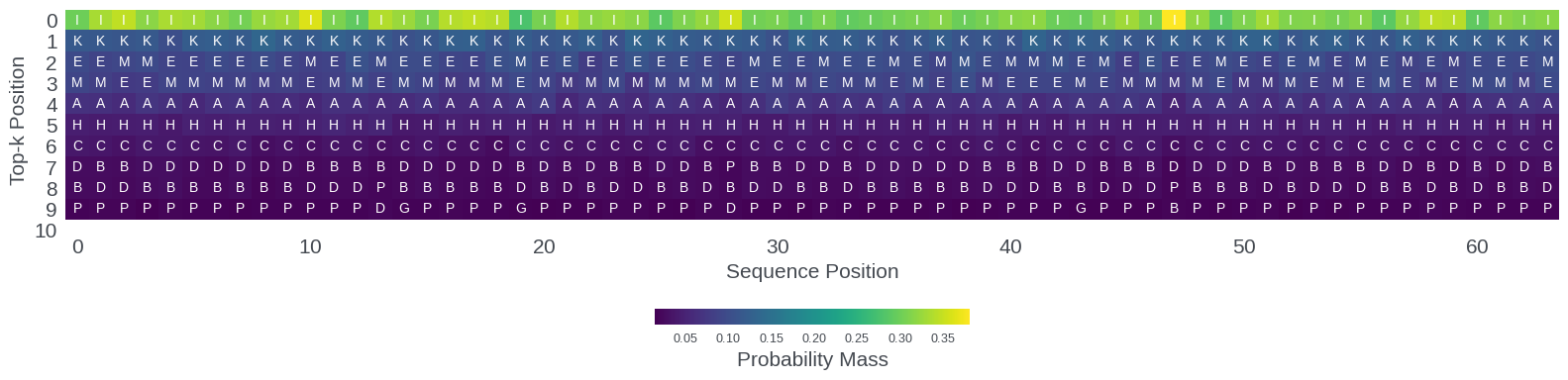}
    \caption{Output layer top-10 token ranking and probability mass distribution across sequence positions. Letters represent token indices and their rank.}
    \label{fig:output_vis}
\end{figure}

\paragraph{The Output Layer and Vocabulary Probability Distribution}
The final, and most consequential, question concerns the impact of positional perturbations on the output distribution of the language modeling head, which serves as the training signal in logit-based KD. Analysis of the output layer (Figure~\ref{fig:output_vis}) reveals that while the top-10 tokens by rank remain largely consistent across positions, both their associated probability mass and their ranking order exhibit position-dependent fluctuations. These subtle but systematic changes demonstrate that the distillation target is not solely a function of the input tokens' semantic and syntactic content; it is also modulated by the positional encoding, even across packed samples within the input sequence.

This finding has a direct and important implication: the teacher model's output distribution encodes position-dependent signals that are transmitted to the student through the distillation loss. This occurs even across sample boundaries within a packed sequence, where individual segments do not attend to one another. The positional perturbations introduced by RoPE are therefore not merely local modifications to query and key states, they are propagated and amplified throughout the entire transformer architecture, ultimately influencing the model's output logits.

This analysis directly answers RQ~(3): positional perturbations do propagate meaningfully through the teacher model's forward pass, and the resulting position-dependent output distributions serve as an implicit training signal for positional information in the student model. Combined with the KD-over-CE gap observed in Section~\ref{sec:ablations}, this provides evidence that KD enables student models to acquire stronger long-context retrieval capabilities than CE training alone, even when both use identical packed short-context data\footnote{Our results demonstrate that \emph{KD enhances} long-context extension on packed short data, rather than that KD is the \emph{sole source} of long-context ability.}.

\section{Hidden State Pattern Evolution During Long-Context Extension}
\label{sec:extension_patterns}

Having established that positional perturbations propagate through the teacher's forward pass and influence the distillation signal (Section~\ref{sec:analysis}), we now investigate how the student model's internal representations change during long-context extension. This analysis directly addresses RQ~(4): can we identify specific patterns in the student model's parameter updates that correlate with positional information?

\subsection{Method: Checkpoint Comparison}

To this end, we compare hidden states produced by two checkpoints: the Phase~1 checkpoint (trained with short context of 2,048 tokens per sequence) and the Phase~2 checkpoint (further trained with extended context of 128,000 tokens per sequence). To isolate the effects of long-context training from other sources of variation, we evaluate both checkpoints on the same input sample using the repeated-token methodology described in Section~\ref{sec:analysis}. This allows us to examine changes along two axes: Hidden-state dimensions and sequence positions, while controlling for semantic and syntactic content.

\subsection{Findings}
\begin{figure}[t]
    \centering
    \includegraphics[width=0.48\linewidth]{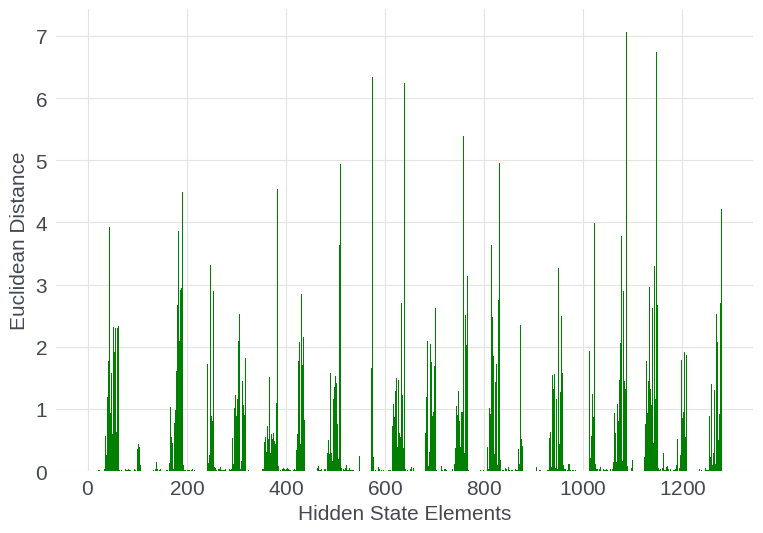}
    \includegraphics[width=0.48\linewidth]{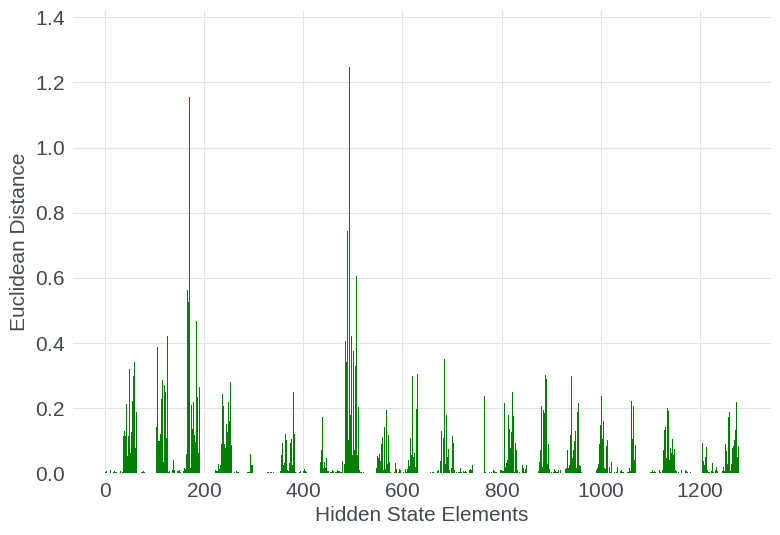}
    \caption{Euclidean distance between Phase~1 and Phase~2 pre-trained hidden states per hidden-state element for Layer~0 (top) and Layer~30 (bottom).}
    \label{fig:phase_analysis_per_dim}
\end{figure}

\begin{figure}[t]
    \centering
    \includegraphics[width=0.48\linewidth]{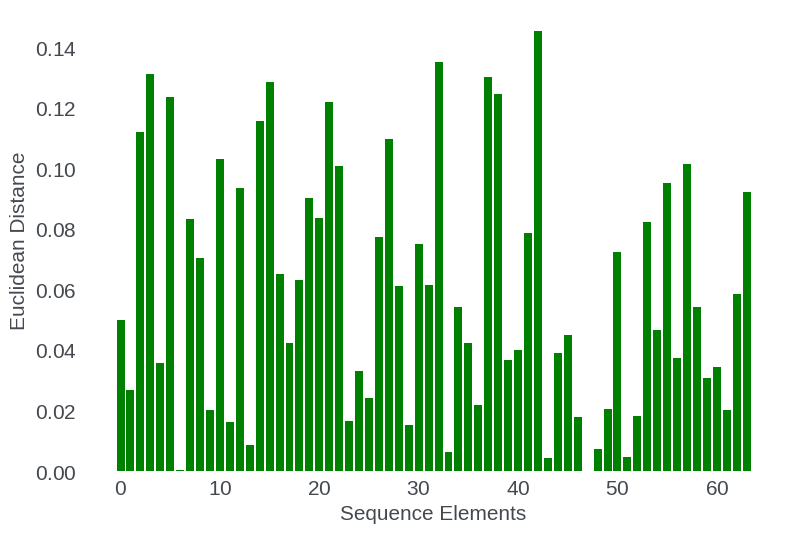}
    \includegraphics[width=0.48\linewidth]{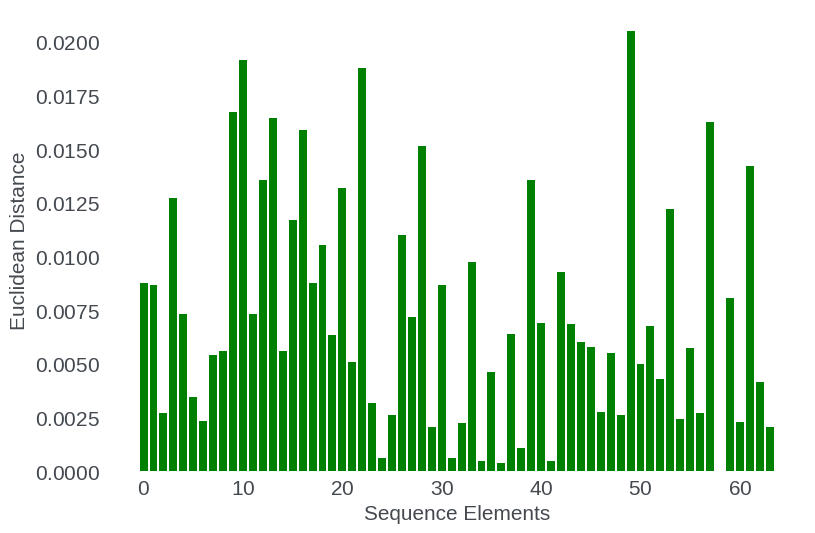}
    \caption{Euclidean distance between Phase~1 and Phase~2 pre-trained hidden states per sequence element for Layer~0 (top) and Layer~30 (bottom).}
    \label{fig:phase_analysis_per_seq}
\end{figure}

\paragraph{Dimension-Specific Adaptation}
Figures~\ref{fig:phase_analysis_per_dim} and~\ref{fig:phase_analysis_per_seq} show the Euclidean distance between query hidden states produced by the Phase~1 and Phase~2 checkpoints. Figure~\ref{fig:phase_analysis_per_dim} reveals a striking pattern: certain dimensions within the hidden-state vector ($\text{dim} = 1{,}280$) exhibit significantly larger differences between training phases than others. Distinct spans of parameters undergo substantial modification during long-context training, while intervening spans remain relatively unchanged. Notably, the most pronounced adaptation occurs in higher-index dimension pairs (corresponding to lower RoPE frequencies, i.e., larger $i$ in Eq.~\ref{eq:theta_frequency}), which are precisely the dimensions responsible for encoding long-range positional relationships as illustrated in Figure~\ref{fig:rope_intuition}. This alignment between the adapted dimensions and the RoPE frequency spectrum suggests that long-context extension selectively refines the rotational components most relevant for distinguishing distant positions.

\paragraph{Position-Independent Adaptation}
When examining the Euclidean distance across sequence positions (Figure~\ref{fig:phase_analysis_per_seq}), we observe that the adaptation magnitude does not meaningfully increase for later positions in the sequence. This indicates that the model's ability to handle long contexts is not achieved by memorizing position-specific representations; rather, it is accomplished by refining dimension-specific positional features that generalize across all positions.

\paragraph{Implications}
Together, these observations suggest that long-context extension does not require wholesale relearning of representations. Instead, it selectively refines a subset of dimensions responsible for encoding long-range positional relationships. This answers RQ~(4) and reinforces our hypothesis about implicit positional distillation: the teacher model's output distribution encodes positional information through RoPE-induced perturbations (as demonstrated in Section~\ref{sec:analysis}), and during long-context extension, this positional signal drives targeted updates to the rotational components in the student model that are most relevant for long-range position encoding.

\section{Related Work}
\subsection{Positional Encoding in Transformers}

Positional encoding is fundamental to transformer architectures~\citep{vaswani2017attention}, with RoPE~\citep{su2024roformer} emerging as the predominant approach in modern LLMs (see Section~\ref{sec:rope_background} for background).

Several works have explored the scaling properties of RoPE. \citet{liu2024scalinglawsropebased} established scaling laws for RoPE-based extrapolation. \citet{peng2024yarn} proposed YaRN, introducing frequency-domain modifications to improve context scaling, while \citet{shang2025longrope2} developed LongRoPE2 for near-lossless context window expansion. \citet{yang2025ropenopeagainnew} explored hybrid strategies combining RoPE with other encoding methods, and \citet{gopalakrishnan2025decouplingwhatwherepolar} offered an alternative geometric perspective through polar coordinate embeddings. Our work differs from these approaches by focusing on how positional information can be \emph{transferred} through knowledge distillation, rather than directly extending positional encoding capabilities.

\subsection{Long-Context Training and Extension}

Extending the context window of language models has been pursued through multiple strategies. Direct long-context pre-training, as implemented in models like Gemma-3~\citep{gemmateam2025gemma3technicalreport} and Qwen-3~\citep{yang2025qwen3technicalreport}, requires substantial computational resources and carefully curated long-document corpora. Continued pre-training approaches fine-tune models on longer sequences after initial training, but often suffer from catastrophic forgetting of short-context capabilities.

Alternative strategies include sparse attention mechanisms that reduce the quadratic complexity of self-attention, memory-augmented architectures that externalize context storage, and sliding window approaches. However, these architectural modifications often require significant changes to the model structure and may not transfer easily across different model families. Our approach is orthogonal to these methods: we demonstrate that long-context capabilities can emerge through the training objective alone, without architectural modifications or long-context training data.

\subsection{Knowledge Distillation for Language Models}

Knowledge distillation (KD), originally proposed for model compression~\citep{hinton2015distilling}, transfers knowledge from a larger teacher to a smaller student model through soft label matching. In language modeling, logit-based KD has proven effective for training compact models that approach the performance of larger counterparts. Recent work on on-device language models has demonstrated that KD can enable student models to acquire capabilities not explicitly present in the training data~\citep{huber2025mobilellmprotechnicalreport,huber2025cosmoescompactsparsemixture}.

\citet{huber2025mobilellmprotechnicalreport} first observed that student models distilled from long-context teachers exhibit long-context abilities even when trained exclusively on short sequences. However, this phenomenon was reported without a mechanistic explanation. Our work provides the first systematic investigation of \emph{why} this transfer occurs, tracing the propagation of positional perturbations through the teacher's forward pass and into the distillation signal. This mechanistic understanding enables principled design choices for RoPE scaling strategies in the knowledge distillation setting.

\section{Conclusion}
This work provides a systematic investigation into how positional information is transferred through knowledge distillation and how model representations evolve during long-context extension in transformer language models using RoPE. Our findings, obtained in the specific setting of a 1.1B student distilled from a Llama-4 Scout teacher using packed short-context data, can be summarized along three axes:

\textbf{Optimal RoPE scaling for KD (RQ~1):} Phase-wise scaling of the RoPE base parameter $\theta$, using a smaller value during short-context pre-training and a larger value during long-context extension, maximizes rotational spectrum utilization at each stage and achieves the best long-context distillation performance. This strategy consistently outperforms both fixed low and fixed high $\theta$ configurations.

\textbf{Implicit positional transfer through KD (RQs~2 and~3):} Long-context retrieval abilities are significantly stronger under KD than under CE training, even when both use exclusively packed short-context data at 128k sequence length. We trace a plausible mechanism: RoPE perturbations applied to query and key vectors propagate through the teacher's forward pass, systematically influencing the output probability distribution. These position-dependent output distributions serve as the distillation signal, carrying implicit positional information to the student model. While the current evidence is correlational, the consistent KD-over-CE gap across all RoPE configurations supports the hypothesis that the teacher's positional signal is a contributing factor.

\textbf{Structured parameter updates during long-context extension (RQ~4):} Model updates during long-context extension are dimension-specific rather than position-specific. Certain spans of hidden-state dimensions exhibit pronounced adaptation, while others remain largely unchanged. No clear monotonic trend is observed across sequence positions, suggesting that long-context generalization is driven by targeted refinement of specific rotational components rather than global shifts in representation.

These insights have practical implications for the design and training of compact and on-device language models in this specific KD setting.

\clearpage

\section{Limitations}
\label{sec:limitations}

We identify several limitations of the current study that should be considered when interpreting our findings:

\textbf{Single teacher-student configuration.} All experiments use a single teacher (Llama-4 Scout, 17B active / 109B total) and a single student (1.1B parameters). The generalization of our findings to other model scales, architectures, or teacher-student pairings remains to be established.

\textbf{Causal ambiguity.} Because Stage~2 trains at 128k context length for both KD and CE conditions, our experimental design does not fully isolate KD as the sole source of long-context ability. The observed KD-over-CE advantage is consistent with our hypothesis, but targeted ablation experiments (e.g., using a short-context teacher or position-invariant distillation targets) are necessary to establish definitive causality.

\textbf{Evaluation scope.} Our long-context evaluations rely on Needle-in-a-Haystack and RULER, both of which are synthetic retrieval-style benchmarks. Performance on additional long-context tasks such as long-document QA, multi-document summarization, or multi-turn dialogue can likely shed further light on the long-context abilities instilled from logit distillation.

\textbf{Correlational mechanistic analysis.} The propagation analysis in Section~\ref{sec:analysis} demonstrates that positional perturbations reach the output layer, but does not include causal interventions (e.g., freezing or ablating specific dimensions) that would further establish the observed transfer.

\bibliographystyle{assets/plainnat}
\bibliography{paper.bib}

\newpage
\appendix
\onecolumn

\section{Experimental Datapoint and Token Sequence}
\label{app:inputs_and_tokens}
\begin{tcolorbox}[colback=blue!5!white, colframe=blue!50!white, title=Randomly Chosen Example to Show Positional Impact, boxrule=0.8mm, arc=3mm, left=2mm, right=2mm, top=1mm, bottom=1mm]
\textless BOS\textgreater Taking cereals for breakfast has become a culture especially in the US. The average person takes cereals at least once a day. The reason might be because they all taste good, but one thing that many people might be aware of is the numerous health benefits that they have. [...]. Therefore, it is evident to claim that cereals have a lot of advantages. Below are some of the known benefits of taking breakfast cereals.\\
Why you and your kids should take \textless EOS\textgreater
\end{tcolorbox}

\begin{tcolorbox}[colback=red!5!white, colframe=red!50!white, title=Resulting Token Sequence, boxrule=0.8mm, arc=3mm, left=2mm, right=2mm, top=1mm, bottom=1mm]
200000, 86256, 23904, 1331, 393, 32044, 947, 7375, 262, 11773, 9924, 310, 290, 2792, 26, 589, 4745, 2084, 7790, 23904, 1331, 552, 5281, 6837, 262, 3364, 26, 589, 6379, 4824, 446, 2895, 1451, 806, 25405, 3049, 24, 1108, 1085, 7744, 511, 2233, 2721, 4824, 446, 21758, 323, 373, 290, 22427, 4500, 12093, 511, 1451, 847, 26, [...], 24, 536, 373, 34727, 328, 7292, 511, 23904, 1331, 847, 262, 5693, 323, 26286, 26, 34635, 583, 1696, 323, 290, 5711, 12093, 323, 9096, 32044, 23904, 1331, 335, 17375, 650, 341, 913, 16767, 2036, 3373, 220, 200001\\
(Repeated 64x to fill seq\_len of 128k tokens.)
\end{tcolorbox}

\begin{tcolorbox}[colback=olive!5!white, colframe=olive!50!white, title=Resulting Analysis Sequence, boxrule=0.8mm, arc=3mm, left=2mm, right=2mm, top=1mm, bottom=1mm]
220, 220, 220, 220, 220, 220, 220, 220, 220, 220, 220, 220, 220, 220, 220, 220, 220, 220, 220, 220, 220, 220, 220, 220, 220, 220, 220, 220, 220, 220, 220, 220, 220, 220, 220, 220, 220, 220, 220, 220, 220, 220, 220, 220, 220, 220, 220, 220, 220, 220, 220, 220, 220, 220, 220, 220, 220, 220, 220, 220, 220, 220, 220, 220\\
(Each with 2,048 tokens in between)
\end{tcolorbox}

\end{document}